\documentclass[10pt,twocolumn,letterpaper]{article}
\usepackage[accsupp]{axessibility} 
\usepackage{iccv}
\usepackage{times}
\usepackage{epsfig}
\usepackage{graphicx}

\usepackage{amsmath}
\usepackage{amssymb}

\usepackage{wrapfig}
\usepackage{enumitem}
\usepackage{multirow}
\usepackage{algorithmic}
\usepackage{algorithm}
\usepackage{diagbox}    
\usepackage{booktabs}
\usepackage{caption}    
\usepackage{subcaption} 
\usepackage{bm}
\usepackage[table]{xcolor}
\definecolor{mygray}{gray}{0.9}
\definecolor{comment}{RGB}{0, 102, 0}

\newcommand{\eqname}[1]{\tag*{#1}}

\usepackage[breaklinks=true,bookmarks=false]{hyperref}

\iccvfinalcopy 

\def\iccvPaperID{9790} 

\ificcvfinal\fi

\begin{document}

\title{Invariant Feature Regularization for Fair Face Recognition}

\author{%
 \textbf{Jiali Ma}\textsuperscript{1} \quad 
 \textbf{Zhongqi Yue}\textsuperscript{2} \quad 
 \textbf{Kagaya Tomoyuki}\textsuperscript{3} \quad 
 \textbf{Suzuki Tomoki}\textsuperscript{3}\\ 
 \textbf{Karlekar Jayashree}\textsuperscript{1} \quad 
\textbf{Sugiri Pranata}\textsuperscript{1} \quad 
 \textbf{Hanwang Zhang}\textsuperscript{2}\\
\small \textsuperscript{1}Panasonic R\&D Center Singapore\quad \textsuperscript{2}Nanyang Technological University\quad 
\textsuperscript{3}Panasonic Connect Co., Ltd. R\&D Division\\
\tt\small jiali.ma@sg.panasonic.com \quad yuez0003@ntu.edu.sg \quad kagaya.tomoyuki@jp.panasonic.com\\
\tt\small suzuki.tomoki@jp.panasonic.com \quad karlekar.jayashree@sg.panasonic.com \\
\tt\small sugiri.pranata@sg.panasonic.com \quad hanwangzhang@ntu.edu.sg\\}


\maketitle
\ificcvfinal\thispagestyle{empty}\fi

\begin{abstract}

Fair face recognition is all about learning invariant feature that generalizes to unseen faces in any demographic group.
Unfortunately, face datasets inevitably capture the imbalanced demographic attributes that are ubiquitous in real-world observations,
and the model learns biased feature that generalizes poorly in the minority group.
We point out that the bias arises due to the confounding demographic attributes,
which mislead the model to capture the spurious demographic-specific feature.
The confounding effect can only be removed by causal intervention, which requires the confounder annotations.
However, such annotations can be prohibitively expensive due to the diversity of the demographic attributes.
To tackle this, we propose to generate diverse data partitions iteratively in an unsupervised fashion. Each data partition acts as a self-annotated confounder,
enabling our Invariant Feature Regularization (\textsc{Inv-Reg}) to deconfound.
\textsc{Inv-Reg} is orthogonal to existing methods, and combining \textsc{Inv-Reg} with two strong baselines (Arcface and CIFP) leads to new state-of-the-art that improves face recognition on a variety of demographic groups. 
Code is available at \small \url{https://github.com/milliema/InvReg}.

\end{abstract}

\section{Introduction}

Face recognition is essentially an out-of-distribution generalization problem, where the goal is to learn feature that generalizes to unseen faces in deployment~\cite{turk1991eigenfaces}. In particular, due to its wide application in sensitive areas such as crime prevention~\cite{woodward2003biometrics}, fair face recognition becomes a pressing need~\cite{drozdowski2020demographic,hupont2019demogpairs,wang2019racial}.
This means that the model must perform equally well on all demographic groups, \ie, it learns the causal feature (\eg, face identity) invariant to the demographic attributes (\eg, race or gender).

%
\begin{figure}[t!]
    \centering 
    \begin{subfigure}[t]{1\linewidth}
         \includegraphics[width=1\linewidth]{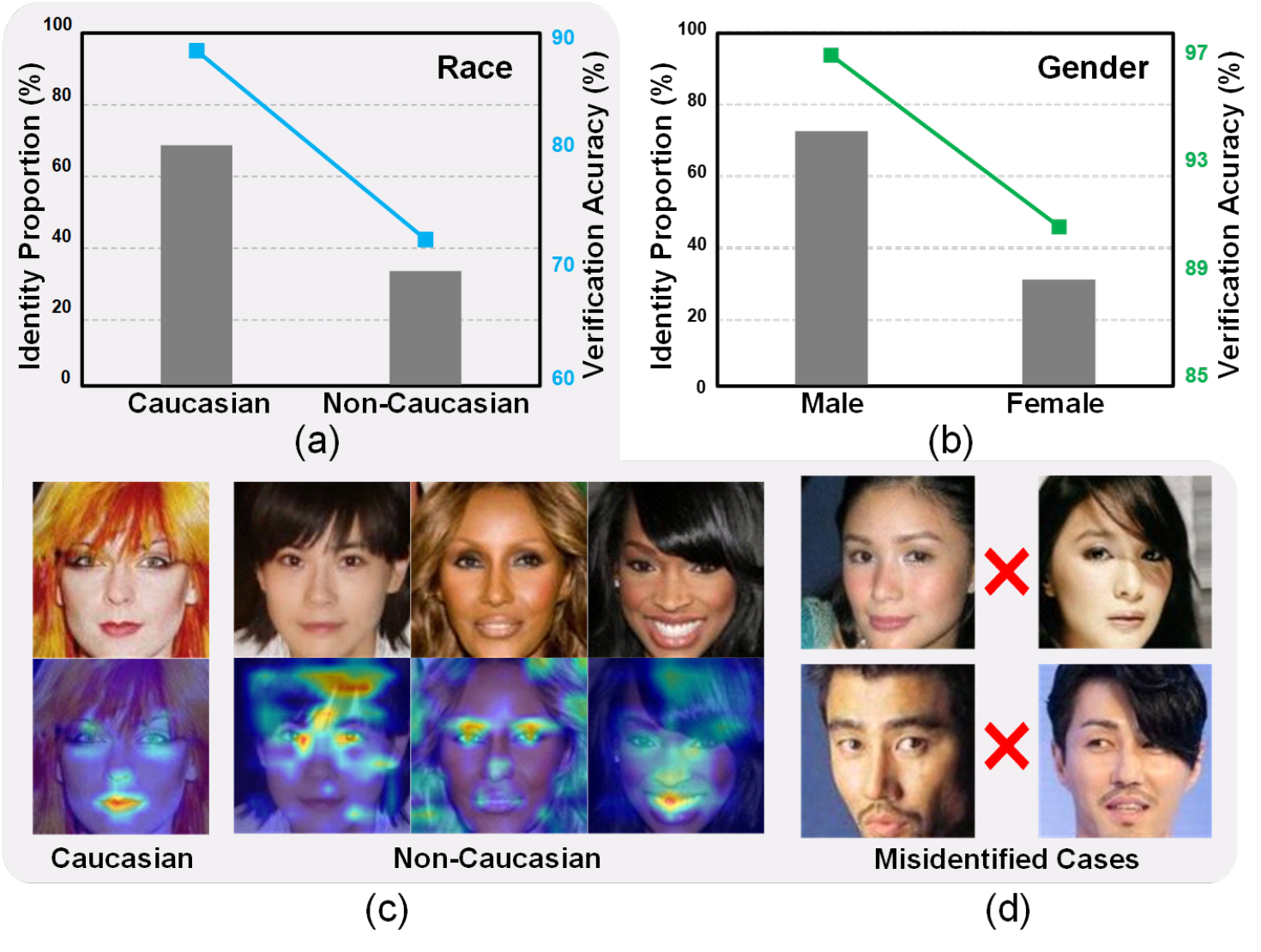}
         \phantomcaption
         \label{fig:1a}
    \end{subfigure}
    \begin{subfigure}[t]{0\linewidth} 
         \includegraphics[width=\linewidth]{example-image-b}
         \phantomcaption
         \label{fig:1b}   
    \end{subfigure}
    \begin{subfigure}[t]{0\linewidth} 
         \includegraphics[width=\linewidth]{example-image-b}
         \phantomcaption
         \label{fig:1c}   
    \end{subfigure}
    \begin{subfigure}[t]{0\linewidth} 
         \includegraphics[width=\linewidth]{example-image-b}
         \phantomcaption
         \label{fig:1d}
    \end{subfigure}
    \vspace*{-11mm}
    \caption{Biased model trained on the dataset with imbalanced demographic attributes. (a) and (b) show the attribute proportions and verification accuracy on race and gender, respectively. (c) Grad-CAM~\cite{selvaraju2017grad} attention maps of face images from two racial groups. (d) Misidentified cases.}
    \label{fig:1}
    \vspace*{-4mm}
\end{figure}
\begin{figure*}[t!]
    \centering 
    \begin{subfigure}[t]{0.9\linewidth}
         \includegraphics[width=\linewidth]{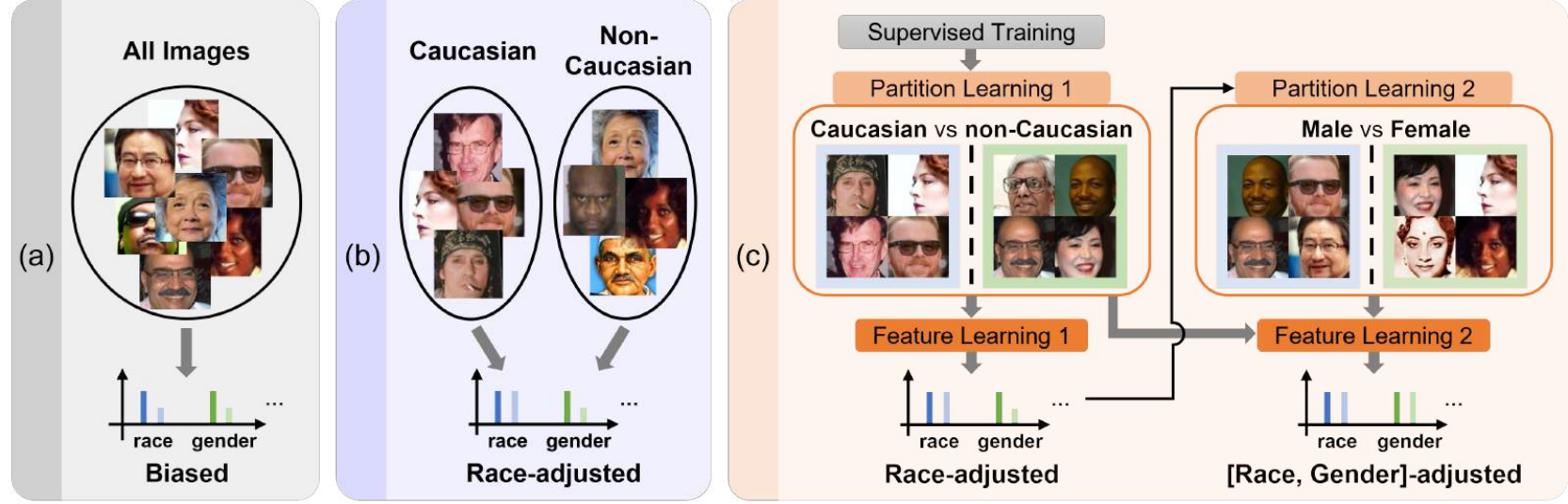}
         \phantomcaption
         \label{fig:2a}
    \end{subfigure}
    \begin{subfigure}[t]{0\linewidth} 
         \includegraphics[width=\linewidth]{example-image-b}
         \phantomcaption
         \label{fig:2b}   
    \end{subfigure}
    \begin{subfigure}[t]{0\linewidth} 
         \includegraphics[width=\linewidth]{example-image-b}
         \phantomcaption
         \label{fig:2c}   
    \end{subfigure}
    \vspace*{-5mm}
    \caption{Comparison of different face recognition approaches. (a) Standard supervised training (biased to race and gender). (b) Learning with ground-truth race partition (biased to gender). (c) Our \textsc{Inv-Reg} without using any annotation of ground-truth demographic attribute (invariant to race and gender).}
    \label{fig:2}
    \vspace*{-4mm}
\end{figure*}
However, demographic attributes are naturally imbalanced in data at scale, \eg, as shown in Figure~\ref{fig:1} first row, in the prevailing MS-Celeb-1M dataset~\cite{guo2016ms}, ``non-Caucasian'' and ``female'' are the minority racial and gender groups, respectively.
Furthermore, a model naively trained on this imbalanced dataset underperforms on the minority groups in deployment, \eg, more than 10\% accuracy decrease on ``non-Caucasian'', and 5\% decrease on ``female''.
In particular, by analyzing the attention maps across the two racial groups in Figure~\ref{fig:1c}, we observe that besides the causal feature (\eg, facial attributes like eyes and nose), the model additionally focuses on the spurious demographic-specific feature (\eg, hairstyle) on the minority group. 
This is because the less diverse face images in each non-Caucasian identity tend to share the hairstyle, making it a valid context to distinguish identities.
Yet this demographic-specific context generalizes poorly to unseen faces, \eg, misidentifying the same identity with different hairstyles as in Figure~\ref{fig:1d}.

From a causal point of view, the bias stems from the confounding effect~\cite{pearl2009causality,pearl2000models, gong2021mitigating}.
In face recognition, when trained to predict the identity label $Y$ given the face image $X$, the model is confounded by the demographic attributes $D$ (\eg, $D=$``non-Caucasian''), which is the common cause of $X$ and $Y$.
Specifically, the training image $X$ is sampled from the demographic group specified by the attributes $D$ (\ie, $D \to X$), and any demographic-specific feature that is discriminative towards $Y$ (\eg, hairstyle) serves as contextual cue when predicting the identity (\ie, $D \to Y$).
Hence, in pursuit of a lower classification loss, the model recklessly exploits the context feature ($X\leftarrow D \to Y$) which may not generalize to unseen faces.
The confounding effect is more pronounced on the minority group, \eg, in the extreme case where $D=$``dark skin'' only has one identity, learning the ``dark skin'' feature alone ($D\to Y$) can tell the identity apart from ``light skin'' ones, yet this spurious feature fails to distinguish different ``dark skin'' identities in deployment.

The aforementioned confounding effect can only be removed by causal intervention~\cite{pearl2000models,pearl2009causality}.
One way is to collect a balanced dataset with diverse faces in all demographic groups (\ie, adjusting the demographic attribute $D$). 
After all, if there is no demographic-specific context in any group, the model can only learn the causal feature. 
However, this is impractical due to the prohibitive data collection cost.
Note that ``balanced'' without ``diverse'' is ineffective since the spurious context still persists, as shown by the limited success on small balanced dataset~\cite{wang2020mitigating}.
The other way is backdoor adjustment~\cite{pearl2009causality}, which can be implemented by data partition (Section~\ref{sec:3.2}). For example, in contrast to the naive supervised training on all images in Figure~\ref{fig:2a}, some works~\cite{gong2021mitigating,wang2020mitigating,liu2022learning,wang2019racial} partition the training images into ``race'' splits (\ie, adjusting ``race''), and train a model invariant across the splits as illustrated in Figure~\ref{fig:2b}. However, since the demographic attributes are diverse in practice, using only the ``race'' splits is far from sufficient.

To this end, we propose Invariant Feature Regularization, dubbed as \textbf{\textsc{Inv-Reg}}, which iteratively self-annotates the confounders by learning data partitions, as illustrated in Figure~\ref{fig:2c}.
Our \textsc{Inv-Reg} hinges on the invariance of causal relation~\cite{pearl2009causality,parascandolo2018learning}:
causal feature is invariantly discriminative across the splits in any confounder partition.
Its contra-position enables \textbf{1) partition learning} about confounder: if the current feature is not invariantly discriminative across the learned splits, the partition corresponds to a confounding demographic attribute. Then we perform \textbf{2) feature learning} to achieve invariance across the learned splits (\ie, causal intervention), which removes its confounding effect. We iterate between the two steps to learn causal feature invariant to diverse demographic attributes.

\noindent Our contributions are summarized below:
\begin{itemize}[leftmargin=+0.1in,itemsep=2pt,topsep=0pt,parsep=0pt]
    \item We propose a partition learning strategy to self-annotate the demographic attributes (\ie, confounder) in the form of data partitions (Section~\ref{sec:4.1}). In particular, our approach discovers diverse demographic partitions without relying on any ground-truth annotation.
    \item We use the discovered partitions to impose an invariant regularization in training to learn causal feature robust in all demographic groups (Section~\ref{sec:4.2}).
    \item Overall, \textsc{Inv-Reg} is a regularization module orthogonal to existing face recognition methods. Combining \textsc{Inv-Reg} with two strong baselines leads to new state-of-the-art results, \ie, 79.44\% on Arcface and 81.17\% on CIFP for average multi-racial accuracy (Section~\ref{sec:5.2}).
\end{itemize}
\section{Related Work}
\noindent\textbf{Fair Face Recognition}. 
Conventional face recognition explores different loss functions for improved feature learning. 
For example, CenterLoss~\cite{wen2016discriminative} and RangeLoss~\cite{zhang2017range} penalize the distance between feature and the corresponding class center.
Sphereface~\cite{liu2017sphereface}, Cosface~\cite{wang2018cosface}, and Arcface~\cite{deng2019Arcface} leverage margin loss to promote intra-class density and inter-class separability.
However, they are easily biased towards the majority group in the training data.
In pursuit of fairness, some works construct balanced datasets \wrt race and gender attributes at limited scale~\cite{wang2020mitigating,hupont2019demogpairs, robinson2020face,wang2019racial}. 
Other works leverage group-level (\eg, race)~\cite{wang2020mitigating,liu2022learning} or sample-level margins~\cite{xu2021consistent,cao2020domain,liu2019fair, dong2017class, wang2020mis, wu2022boundaryface} for re-weighted training, such that minority groups or hard samples (\eg, misclassified training samples) are assigned a larger training weight.
In contrast, our \textsc{Inv-Reg} grounds the elusive confounding demographic attributes into concrete data partitions as self-annotated confounders and performs causal intervention to explicitly remove the bias, leading to state-of-the-art accuracy across a variety of demographic groups.


\noindent\textbf{Partition-Based Learning} is an effective tool towards out-of-distribution generalization. 
Some works ~\cite{sagawa2019distributionally,zhang2020coping} divide the training data into attribute-based groups and optimize the worst-group loss. Other works~\cite{sagawa2020investigation, byrd2019effect,yucer2020exploring} down-weight or sub-sample the majority group to artificially balance the confounding attributes. In face recognition, a few works ~\cite{wang2019racial,wang2020mitigating} partition data by race and transfer knowledge from the majority ``Caucasian" to other races. However, they rely on ground-truth race annotations and lack diversity in practice due to the expensive labeling cost.
Some works also leverage self-annotated partitions, \eg, to improve the attention module~\cite{wang2021causal}, learn a disentangled representation~\cite{wang2021self}, or remove spurious background feature ~\cite{creager2021environment}.
In contrast, our \textsc{Inv-Reg} is tailored for fair face recognition and critically differs in 1) assigning partitions at identity-level instead of image-level to remove spurious demographic attributes defined across identities; 2) performing iterative partition learning and feature learning steps to progressively address various spurious features (\eg, race, gender, \etc).


\section{Problem Formulation}

\subsection{Face Recognition}

\noindent\textbf{Data and Model}. We denote training data as $\{\mathbf{x}_i, y_i\}_{i=1}^N$, where $\mathbf{x}_i$ is an image and $y_i\in \{1,\ldots,C\}$ denotes its identity label among $C$ different identities. We drop the subscript $i$ for simplicity when the context is clear.
%
The demographic attribute labels are generally not available in training.
The model consists of a feature backbone $\Phi(\mathbf{x})$ that outputs a feature vector, and a classifier $f$ that outputs the prediction logit for each of the $C$ identities given a feature.

\noindent\textbf{Train and Test}. In training, $\Phi$ and $f$ are optimized by classification loss (\eg, cross-entropy) for identity prediction.
In testing, $\Phi$ is evaluated on unseen samples $\{\mathbf{x}_i, y_i, \mathbf{d}_i\}_{i=1}^M$ where $\mathbf{d}_i$ is a tuple of demographic attributes, \eg, $\mathbf{d}_i = (\textrm{race: ``Caucasian''}, \textrm{gender: ``Male''})$. A fair face recognition model should learn feature $\Phi(\mathbf{x})$ that is discriminative towards $y$ in each demographic group (\eg, ``Caucasian'', ``Male''). The exact evaluation protocol is in Section~\ref{sec:5.1}.

\begin{wrapfigure}{r}{0.15\textwidth}
    \centering
    \vspace{-2mm}
    \includegraphics[width=0.13\textwidth]{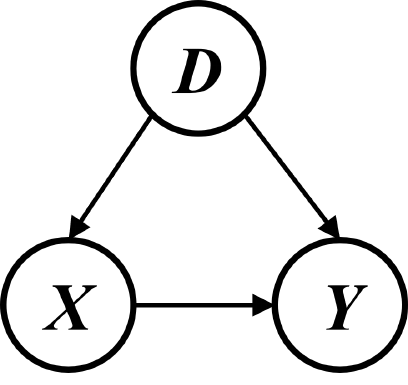} 
    \vspace{-2.5mm}
    \caption{Causal graph of face recognition.}
    \label{fig:fig3}
    \vspace{-1.5mm}
\end{wrapfigure}

\noindent\textbf{Causal View}. We depict the causal relations in face recognition with a causal graph in Figure~\ref{fig:fig3}. $X\to Y$ represents the desired causal effect from an image $X$ to its identity $Y$, 
$D$ denotes the set of \emph{all} demographic attributes and is diverse in practice (\eg, CelebA in Section~\ref{sec:5.2}). 
$D\to X$ denotes that the attributes affect the appearance of image $X$ (\eg, few non-Caucasians have light skin).
In particular, $D\to Y$ is because $D$ contains demographic-specific contextual cue for predicting $Y$ due to dataset bias (\eg, hairstyle is discriminative due to limited non-Caucasian training images in Figure~\ref{fig:1c}).
$X\leftarrow D \to Y$ is known as the backdoor path, which misleads the model to capture the spurious confounding effect~\cite{pearl2009causality,pearl2000models} (\eg, predicting with hairstyle in Figure~\ref{fig:1d}). This confounding effect can only be removed by causal intervention~\cite{pearl2000models}, and we introduce an effective implementation below called invariant learning.

\begin{figure*}[t!]
\centering
\noindent\includegraphics[width=1.0\textwidth]{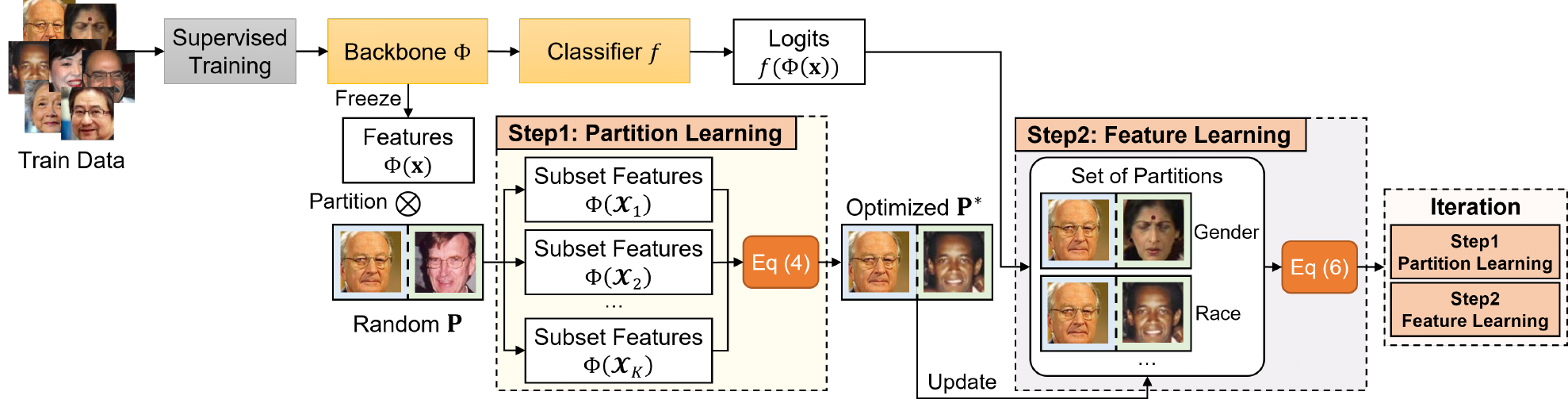} 
\vspace{-5mm}
\caption{Pipeline of \textsc{Inv-Reg}. 
After standard supervised training for 1 epoch, we first conduct [\textbf{Step 1} Partition Learning] to discover a partition corresponding to a confounding demographic attribute (Section~\ref{sec:4.1}). Next, we perform [\textbf{Step 2} Feature Learning] to achieve invariance across the partition subsets (Section~\ref{sec:4.2}). We iterate the above two steps until convergence.}
\label{fig:fig4}
\vspace{-2mm}
\end{figure*}

\subsection{Invariant Learning}
\label{sec:3.2}

Given a partition of the training dataset into subsets and a loss function, the goal is to learn a model that simultaneously minimizes the loss in all the subsets of the partition.

\noindent\textbf{Partition}. In particular, we specify a partition of the $C$ training identities into $K$ subsets by the partition matrix $\mathbf{P}\in \{0,1\}^{C\times K}$, where $P_{y,k}=1$ if the $y$-th identity belongs to the $k$-th subset, and $0$ otherwise.

\noindent\textbf{Loss}. We are interested in supervised classification loss in face recognition.
Given a feature extractor $\Phi$ and classifier $f$, we denote the loss in the $k$-th subset of $\mathbf{P}$ as $\mathcal{L}_{cls}(\Phi, f, \mathbf{P},k)$.

\noindent\textbf{Objective}. Invariant learning uses the following objective, called the invariant risk minimization (IRM)~\cite{arjovsky2019invariant}:
\begin{equation}
    \setlength{\abovedisplayskip}{3pt}
    \setlength{\belowdisplayskip}{3pt}
    \begin{split}\
        &\mathop{\mathrm{min}}_{\Phi,f} \sum_{k=1}^K \mathcal{L}_{cls}(\Phi, f, \mathbf{P},k),\\
        \mathrm{s.t.}\; f \in &\bigcap_{k=1}^K \mathop{\mathrm{argmin}}_{f^*} \mathcal{L}_{cls}(\Phi, f^*, \mathbf{P},k).
        \label{eq:1}
    \end{split}
\end{equation}
The constrained optimization minimizes the combined classification loss (first line) while maintaining an invariant classifier $f$ by keeping the losses in all subsets simultaneously optimal (second line).
This objective learns the causal feature invariant across the subsets.
For example, in Figure~\ref{fig:1c}, while exploiting ``hairstyle'' can reduce $\mathcal{L}_{cls}$ in the minority ``non-Caucasian'', it hurts the loss in ``Caucasian'' as face images of the same identity may present different hairstyles. Therefore, the spurious ``hairstyle'' feature will be removed to satisfy the invariance constraint.

\setlength{\abovedisplayskip}{3pt}\setlength{\belowdisplayskip}{3pt}
To avoid the challenging bi-level optimization in Eq.~\eqref{eq:1}, we use two practical implementations---IRMv1~\cite{arjovsky2019invariant} and REx~\cite{krueger2021out}, which merge the classification loss and the constraint as a single invariant loss, denoted as $\mathcal{L}_{inv}(\Phi,f,\mathbf{P})$:
\begin{flalign}
    \mathcal{L}_{inv}&(\Phi,f,\mathbf{P}) = \sum_{k=1}^K \mathcal{L}_{cls}(\Phi, f, \mathbf{P},k) \eqname{\textbf{IRMv1}~\cite{arjovsky2019invariant}} \\
    &+ \lambda \lVert \nabla_{\mathbf{w}=\mathbf{1.0}} \mathcal{L}_{cls} (f\circ \Phi, \mathbf{w}, \mathbf{P}, k) \rVert_{2}^2
    \label{eq:IRMv1}
\end{flalign}
\begin{flalign}
    \mathcal{L}_{inv}&(\Phi,f,\mathbf{P}) = \sum_{k=1}^K \mathcal{L}_{cls}(\Phi, f, \mathbf{P},k) \eqname{\textbf{REx~\cite{krueger2021out}}} \\
    &+ \lambda \mathrm{Var} \left(\{ \mathcal{L}_{cls}(\Phi, f, \mathbf{P},k)\}_{k=1}^K \right),
    \label{eq:Rex}
\end{flalign}
where $\lambda$ is the invariant penalty strength, and $\mathrm{Var}(\cdot)$ computes the variance of a set of losses.
Intuitively, the second term in IRMv1~\cite{arjovsky2019invariant} treats the logits produced by the composition $f\circ \Phi$ as a feature and computes the optimality of a baseline classifier $\mathbf{w}=\mathbf{1.0}$. Hence IRMv1 achieves invariance by encouraging a fixed baseline to be simultaneously optimal in all subsets.
In contrast, REx minimizes the variance of all subset losses to achieve invariance. We discuss the choice of $\mathcal{L}_{inv}$ and penalty strength $\lambda$ in Section~\ref{sec:5.4}.
\section{Proposed Method}
\label{sec:4}

The overall pipeline of our proposed \textsc{Inv-Reg} is shown in Figure~\ref{fig:fig4} and summarized in Algorithm~\ref{alg:1}.
We first conduct standard supervised training (\eg, using Arcface) for one epoch. Then we iterate between the following steps until convergence:
1) In partition learning, we fix the model and learn a partition matrix $\mathbf{P}$ that \emph{maximizes} $\mathcal{L}_{inv}$ to discover a confounding demographic attribute (Section~\ref{sec:4.1}).
2) We perform feature learning by \emph{minimizing} $\mathcal{L}_{inv}$ to achieve invariance across all discovered partitions (Section~\ref{sec:4.2}).
Note that our implementation of the two steps is simple, \ie, choosing an appropriate form of the classification loss $\mathcal{L}_{cls}$ to compute $\mathcal{L}_{inv}$. We detail the two steps below.

\subsection{Partition Learning}
\label{sec:4.1}

The goal is to find a partition of the training dataset such that the current feature is not invariantly discriminative across the partition subsets.
For example in Figure~\ref{fig:2c}, the race-adjusted model is not invariant across the gender partition, \ie, it is still confounded by the gender attribute, which should be addressed in the subsequent learning.
Specifically, this goal corresponds to finding a partition matrix $\mathbf{P}^\ast$ that maximizes $\mathcal{L}_{inv}$ (\ie, not invariant), while keeping the feature frozen (\ie, current feature)\footnote{We optimize a continuous partition matrix in $\mathbb{R}^{C\times K}$ to enable backpropagation and threshold it to $\{0,1\}^{C\times K}$. Details in Appendix.}:
\begin{equation}
    \begin{aligned}
        \mathbf{P}^\ast = &\mathop{\mathrm{argmax}}_{\mathbf{P}}\mathcal{L}_{inv}(\Phi,f,\mathbf{P}).
    \end{aligned}
    \label{eq:max}
\end{equation}
In particular, when computing $\mathcal{L}_{inv}(\Phi,f,\mathbf{P})$, we adopt the supervised contrastive loss~\cite{khosla2020supervised} as $\mathcal{L}_{cls}$ to evaluate the feature in each subset of $\mathbf{P}$, as formulated below:
\begin{equation}
    \mathcal{L}_{cls}(\Phi, \cdot, \mathbf{P}, k)=\sum_{\mathbf{x} \in \mathcal{X}_k} \sum_{\mathbf{x}^\textrm{+}\in \mathcal{X}_k} -\mathrm{log} \frac{e^{\Phi(\mathbf{x})^\intercal \Phi(\mathbf{x}^\textrm{+})}}
    {\mathop{\sum}_{\mathbf{x}^\textrm{*}\in \mathcal{X}_k}^{\mathbf{x}^\textrm{*}\neq{\mathbf{x}}} e^{\Phi(\mathbf{x})^\intercal \Phi(\mathbf{x}^\textrm{*})}},
    \label{eq:5}
\end{equation}
where $\cdot$ denotes that the classifier $f$ is not used to compute the loss, $\mathcal{X}_k = \left\{ \mathbf{x} | \mathbf{P}_{y, k}=1\right\}$ is the set of training images in the $k$-th subset of $\mathbf{P}$, and $\mathbf{x}^+$ is an image sharing identity with $\mathbf{x}$ in $\mathcal{X}_k$, known as the positive sample in contrastive learning.
Notice that we do not use $f$ in partition learning.
The intuition is that, face recognition hinges on learning invariantly discriminative \emph{feature} towards identity. 
Hence, we seek for confounders (in the form of partitions) that undermine such invariance on \emph{feature} level.
Our adopted supervised contrastive loss measures if the features from the same identity are clustered, and those from different identities are pushed away, making it a favorable choice in this regard.

Overall, we maintain a set of partitions $\mathcal{P}$ that contains all the discovered partitions. $\mathcal{P}$ is initialized as an empty set before training. After finding each $\mathbf{P}^*$ with Eq.~\eqref{eq:max}, we update it with $\mathcal{P}\leftarrow \mathcal{P} \cup \{\mathbf{P}^*\}$ for subsequent feature learning.


\subsection{Feature Learning}
\label{sec:4.2}
The goal is to learn invariant feature that is simultaneously discriminative across the subsets of all the discovered partitions in $\mathcal{P}$.
For example in Figure~\ref{fig:2c}, by learning with the racial and gender partitions, the model becomes invariant to both demographic attributes.
Hence, we optimize the model $\Phi$ and $f$ by minimizing $\mathcal{L}_{inv}$:
\begin{equation}
    \begin{aligned}
        \mathop{\mathrm{min}}_{\Phi, f}\sum_{\mathbf{P} \in \mathcal{P}} \mathcal{L}_{inv}(\Phi, f, \mathbf{P}),
    \end{aligned}
    \label{eq:min}
\end{equation}
where each $\mathbf{P}\in\mathcal{P}$ is frozen.
As the goal is feature learning, we can plug in the classification loss in existing face recognition methods as $\mathcal{L}_{cls}$ to compute $\mathcal{L}_{inv}$.
In this paper, we choose two strong baselines: Arcface~\cite{deng2019Arcface} and CIFP~\cite{xu2021consistent}.
In a nutshell, Arcface increases the decision boundary in angular space such that features of different identities are separated by a user-defined angular margin at least.
CIFP uses an adaptive margin for sample re-weighting, where hard samples (\ie, misclassified in training) are assigned a stringent margin to increase the training weight.
We include the detailed form of the two losses in Appendix.
\renewcommand{\algorithmicforall}{}

\begin{algorithm}[!t]
  \caption{\textsc{Inv-Reg} Training}
  \label{alg:1}
    \begin{algorithmic}[1]
    
    \STATE {\bfseries Input:} training images $\{\mathbf{x}_i, y_i\}_{i=1}^N$, randomly initialized $\Phi$ and $f$, empty set $\mathcal{P}=\left\{\right\}$
    \STATE {\bfseries Output:} trained $\Phi$
    \STATE Train $\Phi$, $f$ on $\{\mathbf{x}_i, y_i\}_{i=1}^N$ with no partition for 1 epoch
    \REPEAT
            \STATE {}\textcolor{comment}{\# \textbf{Step 1}: Partition Learning (Section~\ref{sec:4.1})}
            \STATE {Compute $\mathcal{L}_{cls}(\Phi, \cdot, \mathbf{P}, k) \forall k\in \{1...K\}$ with Eq.~\eqref{eq:5}}
            \STATE {Freeze $\Phi$, $f$, learn $\mathbf{P}^\ast$ with Eq.~\eqref{eq:max}}
            \STATE {{Update $\mathcal{P}\leftarrow \mathcal{P} \cup \{\mathbf{P}^*\}$}}
            \STATE {}\textcolor{comment}{\# \textbf{Step 2}: Feature Learning (Section~\ref{sec:4.2})}
            \STATE {Compute $\mathcal{L}_{cls}(\Phi, f, \mathbf{P}, k), \forall k\in \{1...K\}$}
            \STATE {Freeze $\mathcal{P}$, learn $\Phi$ and $f$ with Eq.~\eqref{eq:min}}
    \UNTIL{convergence}
    \end{algorithmic}
\end{algorithm}

\section{Experiments}
\label{sec:5}
\subsection{Experimental Setting}
\label{sec:5.1}
\noindent\textbf{Training Dataset}. We adopted the refined MS-Celeb-1M dataset ~\cite{guo2016ms, deng2019Arcface}. It is a representative large-scale dataset consisting of 85K celebrity identities with 5.8M images. It exhibits demographic imbalance, where ``Caucasian'' and ``Male'' are the majority groups as shown in Figure~\ref{fig:1}. Pretrained attribute classification models are utilized for the demographic statistical analysis, details are in Appendix.

\noindent\textbf{Testing Dataset}. We performed extensive experiments on various benchmarks with different emphases, summarized in Table~\ref{tab:dataset}.
We evaluated fair face recognition on 3 datasets with attribute labels:
(1) {\bf MFR}~\cite{deng2021masked} is a large-scale non-celebrity dataset with racial attribute annotations. It serves as a challenging benchmark to test model generalization in different racial groups.
(2) {\bf RFW}~\cite{wang2019racial} is a race-balanced dataset constructed from the original MS-Celeb-1M~\cite{guo2016ms}. We followed ~\cite{liu2022learning} to exclude the overlapping identities from RFW with the training dataset.
(3) {\bf CelebA}~\cite{liu2015faceattributes} provides ground-truth labels for fine-grained attributes \eg, gender, hair color and style. We combined data from validation and test splits to evaluate our models. 
In addition, we used 2 datasets to evaluate conventional face recognition.
(4) {\bf IJB-B} and {\bf IJB-C}~\cite{whitelam2017iarpa, maze2018iarpa} are large-scale datasets with images and continuous video frames collected under unconstrained scenarios.
(5) {\bf DigiFace-1M}~\cite{bae2023digiface} is the latest synthetic dataset of realistic face images. It renders each identity under large variations, \eg, hair density, makeup, face-wear, and expression. Hence we adopted it to challenge the generalization ability of models. We used the first 1K identities for testing, where each identity has 72 images.


\noindent\textbf{Implementation Details}. We used image crops with the size of 112$\times$112 as per standard~\cite{liu2022anchorface}. We adopted the modified ResNet-50 and ResNet-100 in ~\cite{deng2019Arcface} as the backbones.
We trained the models with SGD, using batch size of 512 and 21 total epochs. 
All the experiments were conducted on 4 NVIDIA Tesla V100 GPU with Pytorch framework.
For the evaluation protocol, we reported the True Positive Rate (TPR) at a fixed False Positive Rate (FPR). 
We used FPR of 1e-4 for both IJB-B and IJB-C datasets and followed common settings for the rest if not specifically mentioned. 

\begin{table}[t]
\centering
\scalebox{0.88}{
\setlength\tabcolsep{1.5pt}
\def\arraystretch{1.1}
\begin{tabular}{p{2.8cm}<{\centering}p{3.2cm}<{\centering}p{1.3cm}<{\centering}p{1.3cm}<{\centering}}
\hline\hline
Dataset                  & Description           & {\#Identity}   & {\#Image} \\
\hline
MS-Celeb-1M~\cite{guo2016ms}     & Train set, imbalanced         & 85K          & 5.8M   \\
\hline
MFR~\cite{deng2021masked} & Multi-race, large-scale          & 242K         & 1.6M   \\
\hline
RFW~\cite{wang2019racial}      & Multi-race                              & 11K   & 40K  \\
\hline
CelebA~\cite{liu2015faceattributes}   & 40 fine-grain attributes   & 40K      & 1985 \\
\hline
IJB-B~\cite{whitelam2017iarpa}  & \multirow{2}*{Images, video frames}  & 1845   & 76.8K  \\
IJB-C~\cite{maze2018iarpa}      & ~                                & 3531   & 148.8K  \\
\hline
DigiFace-1M~\cite{bae2023digiface}    & \multicolumn{1}{m{3.2cm}}{Large variations, synthetic faces}          & 1000      & 72K   \\
\hline\hline
\end{tabular}}
\vspace{-1mm}
\caption{Statistics of train and test data.}
\label{tab:dataset}
\vspace{-4mm}
\end{table}


\subsection{Fair Face Recognition}
\label{sec:5.2}
\begin{table*}[h]
\centering
\scalebox{0.9}{
\setlength\tabcolsep{1.5pt}
\def\arraystretch{1.1}
\begin{tabular}{p{0.6cm}<{\centering}p{2.5cm}<{\centering}p{2.1cm}<{\centering}p{2.5cm}<{\centering}p{2.6cm}<{\centering}p{2.6cm}<{\centering}p{1.4cm}<{\centering}p{1.4cm}<{\centering}p{1.5cm}<{\centering}}
\hline\hline
\multicolumn{2}{c}{Method}            & African (AF) & Caucasian (CA) & South Asian (SA) & East Asian (EA) & Avg    & Std    & All \\
\hline
{\multirow{4}{*}{\rotatebox{90}{ResNet-50}}}  &Arcface*~\cite{deng2019Arcface}     & 74.54  & 84.43    & 81.47      & 53.27      & 73.43  & 12.18   & 77.91  \\
~&Ours-Arcface          & \cellcolor{mygray}{77.00}  & \cellcolor{mygray}{85.30}    & \cellcolor{mygray}{82.88}      & \cellcolor{mygray}{54.93}      & \cellcolor{mygray}{75.03}  & \cellcolor{mygray}{11.99}        & \cellcolor{mygray}{78.98}  \\ 
~&CIFP*~\cite{xu2021consistent}       & 77.26  & 85.52    & 83.76      & 55.74      & 75.57  & 11.86      & 80.34  \\
~&Ours-CIFP                         & \cellcolor{mygray}\textbf{79.41}  & \cellcolor{mygray}\textbf{86.53}    & \cellcolor{mygray}\textbf{84.99}      & \cellcolor{mygray}\textbf{57.82}      & \cellcolor{mygray}\textbf{77.19}  & \cellcolor{mygray}\textbf{11.49}     & \cellcolor{mygray}\textbf{81.37}  \\
\hline
{\multirow{5}{*}{\rotatebox{90}{ResNet-100}}} &Anchorface~\cite{liu2022anchorface}& 79.31  & 87.00    & 85.59      & 59.70      & 77.90  & \textbf{10.90}   & 82.06  \\ 
~&Arcface$^\dagger$~\cite{deng2019Arcface}     & 79.12  & 87.18    & 85.50      & 55.81      & 76.90  & 12.54     & 80.73  \\
~&Ours-Arcface           & \cellcolor{mygray}{81.76}  & \cellcolor{mygray}{89.16}    & \cellcolor{mygray}{87.64}      & \cellcolor{mygray}{59.20}      & \cellcolor{mygray}{79.44}  & \cellcolor{mygray}{12.01}      & \cellcolor{mygray}{82.85}  \\ 
~&CIFP*~\cite{xu2021consistent}      & 82.55  & 89.40    & 88.77      & 60.58      & 80.32  & 11.71     & 84.36  \\
~&Ours-CIFP                      & \cellcolor{mygray}\textbf{83.70}  & \cellcolor{mygray}\textbf{90.04}    & \cellcolor{mygray}\textbf{89.01}      & \cellcolor{mygray}\textbf{61.91}      & \cellcolor{mygray}\textbf{81.17}  & \cellcolor{mygray}{11.38}       & \cellcolor{mygray}\textbf{84.73}  \\
\hline \hline
\end{tabular}}
\vspace{-1mm}
\caption{Verification performance (\%) on MFR dataset. (“*”: self-implemented results based on the officially released code. “$^\dagger$”: tested results using the released model from the author. ``Ours-": our results achieved by plugging our \textsc{Inv-Reg} into other baselines. ``Avg"/``Std": average/standard deviation of the accuracy on four races. ``All": accuracy on all the samples.)}
\label{tab:MFR}
\vspace{-1.5mm}
\end{table*}

\begin{table}[t]
\centering
\scalebox{0.85}{
\setlength\tabcolsep{0.05pt}
\def\arraystretch{1.1}
\begin{tabular}{p{3.5cm}<{\centering}p{1.0cm}<{\centering}p{1.0cm}<{\centering}p{1.0cm}<{\centering}p{1.0cm}<{\centering}p{1.0cm}<{\centering}p{1.0cm}<{\centering}}
\hline\hline
Method                               & AF              & CA             & SA             & EA             & Avg            & Std  \\
\hline
Arcface~\cite{deng2019Arcface}       & 97.48           & 98.80          & 97.38          & 96.80          & 97.61          & 0.73 \\
LDAM-Cosface~\cite{cao2019learning}  & 97.80           & 98.93          & 97.50          & 97.23          & 97.86          & 0.65 \\
MetaCW~\cite{jamal2020rethinking}    & 97.86           & 99.13          & 98.11          & 97.73          & 98.20          & 0.55 \\
MvCoM-URFace~\cite{liu2022learning}  & 97.18           & 98.85          & 96.98          & 97.15          & 97.54          & 0.76 \\
MvCoM-Cosface~\cite{liu2022learning} & 98.06           & 99.16          & 98.28          & 97.78          & 98.32          & 0.51 \\
Ours-Arcface          & \cellcolor{mygray}\textbf{98.56} & \cellcolor{mygray}\textbf{99.47}   & \cellcolor{mygray}\textbf{98.76} & 
\cellcolor{mygray}\textbf{98.39} & \cellcolor{mygray}\textbf{98.79} & \cellcolor{mygray}\textbf{0.41} \\
\hline \hline
\end{tabular}}
\vspace{-1.5mm}
\caption{Verification accuracy (\%) on RFW (ResNet-100).}
\label{tab:rfw}
\vspace{-0mm}
\end{table}

\noindent\textbf{MFR}. In Table~\ref{tab:MFR}, we observe clear improvements on \emph{all} races by plugging our \textsc{Inv-Reg} into Arcface and CIFP.
For example on ResNet-50, we outperform Arcface baseline by 2.46\% on African, 1.41\% on South Asian, and 1.66\% on East Asian, and we even improve the majority race by 0.87\% on Caucasian.
Furthermore, our method reduces the standard deviation over both baselines, demonstrating more balanced performance across the demographic groups. Note that the low standard deviation of Anchorface on ResNet-100 is because it sacrifices the accuracy of Caucasian.
Overall, we have the following observations:
1) \textsc{Inv-Reg} greatly enhances the performance of minority groups without sacrificing that of the majority one, leading to the best ``Avg" and ``All" performance. This validates the effectiveness of our invariant learning in removing the confounding effects and capturing causal feature.
2) Our improvement on CIFP is not as significant as that on Arcface. We postulate that this is because CIFP leverages sample re-weighting strategy, which is an approximation to causal intervention~\cite{pearl2000models} and has some effects in removing the confounding bias. Nevertheless, our \textsc{Inv-Reg} further removes the bias and achieves state-of-the-art performance.

\noindent\textbf{RFW}.
In Table~\ref{tab:rfw}, our method achieves the top performance across different races with improved average accuracy and lower standard deviation. In particular, our \textsc{Inv-Reg} does not require demographic annotations, unlike, \eg, ~\cite{liu2022learning}.
Note that the accuracy on RFW is near-saturated since it is drawn from the same distribution as the training data.

\begin{figure*}[t!]
\vspace{-1mm}
\centering
\noindent\includegraphics[width=0.9\textwidth]{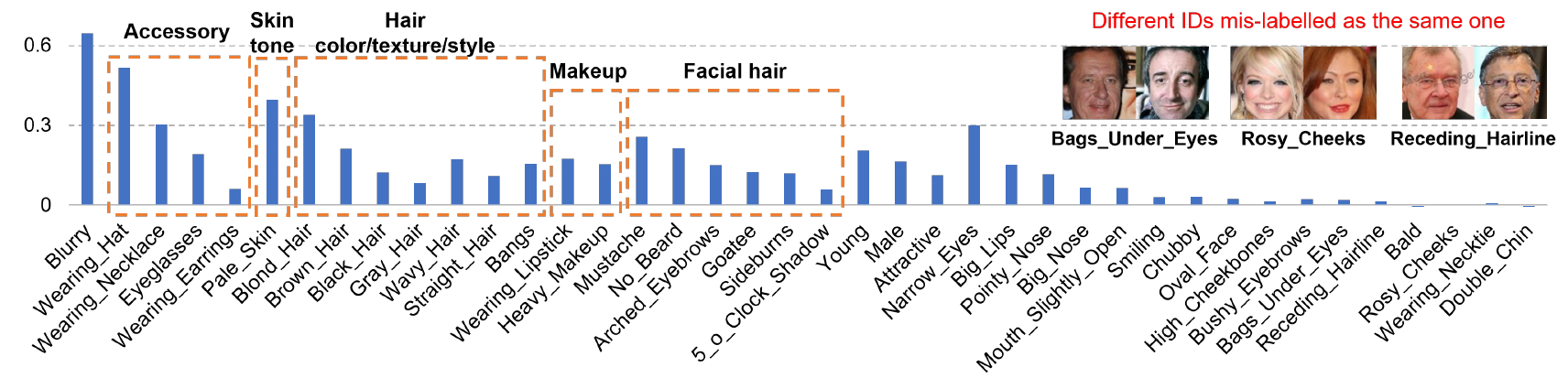} 
\vspace*{-3.5mm}
\caption{Performance improvement (\%) of \textsc{Inv-Reg} over Arcface on fine-grained attributes of CelebA dataset.}
\label{fig:celeb}
\vspace{-2mm}
\end{figure*}
\begin{table}[t]
\centering
\scalebox{0.95}{
\setlength\tabcolsep{0.5pt}
\def\arraystretch{1.1}
\begin{tabular}{p{1.9cm}<{\centering}p{1.6cm}<{\centering}p{1.6cm}<{\centering}p{1.6cm}<{\centering}p{1.6cm}<{\centering}}
\hline\hline
\multirow{2}*{Method}    & \multicolumn{2}{c}{Male}           & \multicolumn{2}{c}{Female} \\ \cmidrule(lr){2-3}\cmidrule(lr){4-5} 
~ & FPR=1e-5  & FPR=1e-4  & FPR=1e-5  & FPR=1e-4  \\  
\hline
Arcface~\cite{deng2019Arcface} & 96.35 & 97.14 & 90.86 & 93.92 \\
Ours    & \cellcolor{mygray}~\textbf{96.61} & \cellcolor{mygray}~\textbf{97.30} & \cellcolor{mygray}~\textbf{91.66} & \cellcolor{mygray}~\textbf{94.17} \\
\hline\hline
\end{tabular}}
\vspace{-1mm}
\caption{Accuracy (\%) on CelebA for different genders.}
\label{tab:gender}
\vspace{-3mm}
\end{table}

\noindent\textbf{CelebA}.
In Figure~\ref{fig:celeb}, we show the improvements by adding \textsc{Inv-Reg} to Arcface (ResNet-100) on each fine-grained attribute.
Our model mitigates racial bias, as evidenced by significant improvements in race-related attributes, \eg, ``Pale\_Skin", ``Blond\_Hair" and ``Black\_Hair".
Furthermore, we effectively eliminate gender bias, \eg, improving gender-related attributes such as ``Wearing\_Lipstick", ``Heavy\_Makeup" and ``Mustache". This is validated by the superior accuracy on Male and Female in Table~\ref{tab:gender}.
Besides race and gender, our method exhibits robustness across a wide range of attributes, such as accessory, facial hair, hair texture and style.
Note that in some attributes our method shows marginal improvements. Possible reasons include:
1) Some attributes are common among a large number of identities in training (\ie, balanced), thus even the baseline captures no bias (\eg, ``Smiling").
2) Some attributes exhibit noisy labels, where faces of different identities are mislabeled as the same one, and we show examples under 3 attributes in Figure~\ref{fig:celeb}.

 \begin{table}[t]
\centering
\scalebox{0.9}{
\setlength\tabcolsep{1.3pt}
\def\arraystretch{1.1}
\begin{tabular}{p{0.5cm}<{\centering}p{4.7cm}<{\centering}p{1.35cm}<{\centering}p{1.35cm}<{\centering}}
\hline\hline
\multicolumn{2}{c}{Method}             & IJB-B       & IJB-C\\
\hline
{\multirow{4}{*}{\rotatebox{90}{ResNet-50}}}  & Arcface~\cite{deng2019Arcface}   & 94.33          & 95.86  \\
~&Ours-Arcface       & \cellcolor{mygray}{94.44}          & \cellcolor{mygray}\textbf{96.15}  \\
~&CIFP~\cite{xu2021consistent}      & 94.57          & 95.80  \\
~&Ours-CIFP          & \cellcolor{mygray}\textbf{94.69} & \cellcolor{mygray}\textbf{96.15} \\
\hline
{\multirow{6}{*}{\rotatebox{90}{ResNet-100}}}  
~&Anchorface-Arcface~\cite{liu2022anchorface}         & 94.42       & 96.22  \\
~&Anchorface-Curricularface~\cite{liu2022anchorface}  & 94.97       & 96.32  \\ \cmidrule(lr){2-4}
~& Arcface~\cite{deng2019Arcface}   & 94.20          & 95.60  \\
~&Ours-Arcface           & \cellcolor{mygray}{94.94}       & \cellcolor{mygray}{96.46}  \\
~&CIFP~\cite{xu2021consistent}      & 95.00          & 96.47  \\
~&Ours-CIFP              & \cellcolor{mygray}\textbf{95.11}       & \cellcolor{mygray}\textbf{96.58} \\
\hline \hline
\end{tabular}}
\vspace{-1mm}
\caption{Verification accuracy (\%) on IJB-B and IJB-C.}
\label{tab:ijb}
\vspace{-3mm}
\end{table}

\subsection{Conventional Face Recognition}
\label{sec:5.3}

\noindent\textbf{IJB-B and IJB-C}.
In Table~\ref{tab:ijb}, our method demonstrates superiority over the baseline models on both test sets regardless of the backbone choice. In particular, combining our \textsc{Inv-Reg} with CIFP leads to the highest performance, showcasing the effectiveness of our proposed method.

\noindent\textbf{DigiFace-1M}.
DigiFace-1M has far greater diversity for each identity and exhibits a domain shift in appearance with the training data. Hence it poses great challenges for model generalization. As shown in Table~\ref{tab:digiface}, our method consistently outperforms both baselines by a substantial margin. This is strong proof that our \textsc{Inv-Reg} captures the causal feature invariant to domain shift and large variations.
\begin{table}[t]
\centering
\scalebox{0.95}{
\setlength\tabcolsep{1.3pt}
\def\arraystretch{1.1}
\begin{tabular}{p{0.6cm}<{\centering}p{2.5cm}<{\centering}p{2cm}<{\centering}p{2cm}<{\centering}}
\hline\hline
\multicolumn{2}{c}{Method}             & FPR=1e-5       & FPR=1e-4\\
\hline
{\multirow{4}{*}{\rotatebox{90}{ResNet-50}}}  & Arcface~\cite{deng2019Arcface}   & 47.03          & 62.71  \\
~&Ours-Arcface       & \cellcolor{mygray}{49.87}          & \cellcolor{mygray}{64.94}  \\
~&CIFP~\cite{xu2021consistent}      & 49.36          & 64.53  \\
~&Ours-CIFP          & \cellcolor{mygray}\textbf{51.19} & \cellcolor{mygray}\textbf{66.06} \\
\hline
{\multirow{4}{*}{\rotatebox{90}{ResNet-100}}}  &Arcface~\cite{deng2019Arcface}   & 48.19          & 64.32  \\
~&Ours-Arcface       & \cellcolor{mygray}{51.80}          & \cellcolor{mygray}{67.57}  \\
~&CIFP~\cite{xu2021consistent}      & 51.73          & 67.08  \\
~&Ours-CIFP          & \cellcolor{mygray}\textbf{53.16} & \cellcolor{mygray}\textbf{68.49} \\
\hline \hline
\end{tabular}}
\vspace{-1mm}
\caption{Verification accuracy (\%) on DigiFace-1M dataset.}
\label{tab:digiface}
\vspace{-2mm}
\end{table}

\subsection{Ablation Study}
\label{sec:5.4}
\begin{figure}[t!]
    \centering 
    \includegraphics[width=0.75\linewidth]{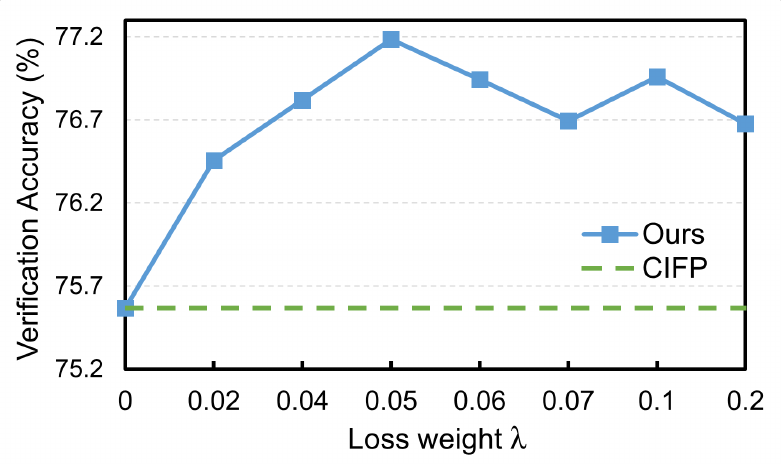} 
    \vspace{-2mm}
    \caption{Accuracy (\%) on MFR dataset with different $\lambda$.}
    \label{fig:lossweight}
    \vspace{-2mm}
\end{figure}

\noindent\textbf{Choice of $\mathcal{L}_{inv}$ and Loss Weight $\lambda$}.
We implemented $\mathcal{L}_{inv}$ with REx in Eq.~\eqref{eq:Rex} for partition learning and with IRMv1 in Eq.~\eqref{eq:IRMv1} for feature learning, which leads to the best performance.
In practice, we find REx is easier to optimize, hence more suitable for the hard task of learning $\mathbf{P}^*$ in Eq.~\eqref{eq:max}, while IRMv1 leads to superior performance in feature learning with its theoretical guarantee~\cite{arjovsky2019invariant}.
We leave the ablation on $\mathcal{L}_{inv}$ in Appendix due to space constraints.
The effect of loss weight $\lambda$ in $\mathcal{L}_{inv}$ is shown in Figure~\ref{fig:lossweight}, where the performance is stable in the range of 0.02 to 0.2 with ResNet-50. Note that it is a standard practice in invariant learning to use a small loss weight, hence $\lambda$ is easy to choose. We adopted the best performing $\lambda=0.05$ in our experiments.

\noindent\textbf{Learned Partition \textit{vs.} Ground-truth One}.
To validate the effectiveness of partition learning, we replaced the learned partitions with the ground-truth race and gender partitions of the training images and followed the same procedure in feature learning. The comparison results with ResNet-50 backbone are in Table ~\ref{tab:explicit_partition}.
Learning with the ground-truth partition leads to improved performance compared to Arcface baseline, validating the effectiveness of invariant learning.
Furthermore, our learned partition achieves the best performance. We postulate the reason is that the feature-level confounders are complex and elusive, which are not fully captured by the limited attribute annotations on the image level. Our \textsc{Inv-Reg} discovers diverse feature-level confounders, allowing it to outperform.

\begin{table}[t]
\centering
\scalebox{0.9}{
\setlength\tabcolsep{0.25pt}
\def\arraystretch{1.1}
\begin{tabular}{p{1.5cm}<{\centering}p{1.2cm}<{\centering}p{1.2cm}<{\centering}p{1.2cm}<{\centering}p{1.2cm}<{\centering}p{1.2cm}<{\centering}p{1.2cm}<{\centering}}
\hline\hline
Method      & AF & CA & SA & EA & Avg  & Std  \\
\hline
Arcface     & 74.54  & 84.43    & 81.47      & 53.27      & 73.43  & 12.18\\
Ours$^\dagger$       & 76.55  & 85.18    & \textbf{83.00}      & 54.61      & 74.84  & 12.10\\
Ours        & \textbf{77.00}  & \textbf{85.30}    & 82.88      & \textbf{54.93}      & \textbf{75.03}  & \textbf{11.99}\\
\hline \hline
\end{tabular}}
\vspace{-1mm}
\caption{Accuracy (\%) on MFR dataset with ground-truth partition (Ours$^\dagger$) and learned partition (Ours).}
\label{tab:explicit_partition}
\end{table}

\begin{table}[t]
\centering
\scalebox{0.9}{
\setlength\tabcolsep{0.25pt}
\def\arraystretch{1.1}
\begin{tabular}{p{1.5cm}<{\centering}p{1.2cm}<{\centering}p{1.2cm}<{\centering}p{1.2cm}<{\centering}p{1.2cm}<{\centering}p{1.2cm}<{\centering}p{1.2cm}<{\centering}}
\hline\hline
Method      & AF & CA & SA & EA & Avg  & Std  \\
\hline
$K=2$  & \textbf{79.41}  & \textbf{86.53} & 84.99          & \textbf{57.82} & \textbf{77.19}  & \textbf{11.49}\\
$K=3$  & 78.97           & 86.39          & 84.88          & 56.67          & 76.73           & 11.91\\
$K=4$  & 79.17           & 86.41          & \textbf{85.11} & 57.07          & 76.94           & 11.79\\
\hline \hline
\end{tabular}}
\vspace{-1mm}
\caption{Accuracy (\%) on MFR dataset with different $K$.}
\label{tab:ablation_subset}
\end{table}

\begin{table}[t]
\centering
\scalebox{0.9}{
\setlength\tabcolsep{0.25pt}
\def\arraystretch{1.1}
\begin{tabular}{p{1.5cm}<{\centering}p{1.2cm}<{\centering}p{1.2cm}<{\centering}p{1.2cm}<{\centering}p{1.2cm}<{\centering}p{1.2cm}<{\centering}p{1.2cm}<{\centering}}
\hline\hline
{\#Partitions}      & AF & CA & SA & EA & Avg  & Std  \\
\hline
1 & 78.61          & \textbf{87.15} & 84.92          & 56.33          & 76.75          & 12.20 \\
2 & 79.24          & 86.44          & \textbf{84.99} & 56.91          & 76.89          & 11.85 \\
3 & \textbf{79.41} & 86.53          & \textbf{84.99} & \textbf{57.82} & \textbf{77.19} & \textbf{11.49} \\
4 & 79.25          & 86.38          & 84.90          & 56.73          & 76.82          & 11.90 \\
\hline \hline
\end{tabular}}
\vspace{-1mm}
\caption{Accuracy (\%) on MFR with different \#partitions.}
\label{tab:ablation_partition}
\vspace{-3mm}
\end{table}

\noindent\textbf{\#Partition Subsets $K$}. By fixing the optimal setting of $\lambda=0.05$, we performed ablations on $K$ with ResNet-50 backbone in Table~\ref{tab:ablation_subset}.
The average accuracy remains stable when we increase $K$ from 2 to 4, with $K=2$ achieving the overall best performance.
Our conjecture is that increasing $K$ makes the optimization more difficult, as the model strives to achieve invariance across all subsets. We will explore an improved optimization strategy in future work.

\noindent\textbf{\#Partitions in $\mathcal{P}$}. In Table~\ref{tab:ablation_partition}, we observe that discovering more partitions generally improves the performance, as the model becomes invariant to more self-annotated confounders. 
However, using more partitions may require larger \#training epochs to fully converge, as evidenced by a slight performance drop with 4 partitions (see Appendix for verification).
Hence we adopted 3 partitions to balance the performance and \#epochs (all models are trained with 21 epochs).

\subsection{Qualitative Analysis}
\label{sec:5.5}

\begin{figure}[t]
    \centering 
    \includegraphics[width=0.95\linewidth]{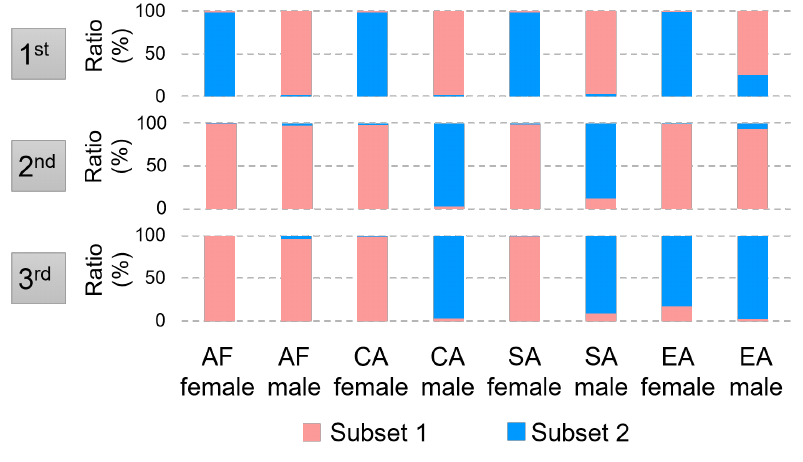} 
    \vspace{-3mm}
    \caption{Proportion of the demographic groups in the subsets of each partition (ResNet-50 backbone).}
    \label{fig:partition}
    \vspace{-0mm}
\end{figure}
\begin{figure}[t]
    \centering 
    \setlength{\abovecaptionskip}{8pt}
    \includegraphics[width=0.9\linewidth]{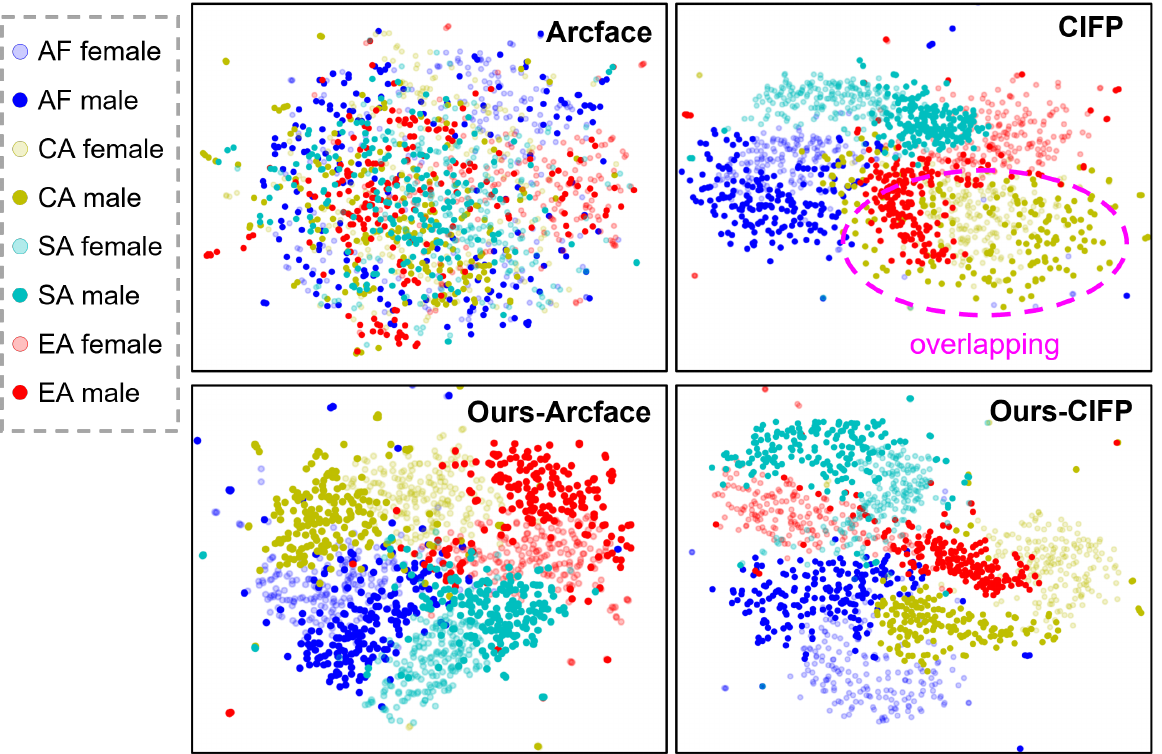} 
    \caption{UMAP visualization~\cite{mcinnes2018umap} of features in different methods and demographic groups (ResNet-50 backbone).}
    \label{fig:umap}
    \vspace{-4mm}
\end{figure}

\noindent\textbf{What does the learned partition capture?}
We visualized the assignment of each demographic group to the 2 subsets of the 3 learned partitions in Figure~\ref{fig:partition}.
We observe that each partition captures a demographic attribute.
The $1^{st}$ partition is about gender, with females and males clearly separated.
In the $2^{nd}$ partition, Caucasian and South Asian males are grouped together, possibly because they share common thick facial hair and sharp facial feature (\eg, high cheekbone and prominent nose).
The $3^{rd}$ partition could be about hair texture, \eg, curly or wavy hair is prevalent in subset 1).
As mentioned earlier, the feature-level confounders captured by the learned partitions can be elusive. Nevertheless, Table~\ref{tab:ablation_partition} shows that learned partitions are more effective than those based on the ground-truth attributes.


\noindent\textbf{What does the invariant feature look like?} We visualized the features of 1,600 identities randomly selected from four races and both genders in Figure~\ref{fig:umap}.
We observe that compared to Arcface baseline, CIFP has some effects in clustering the features of demographic groups, while our \textsc{Inv-Reg} has the best clustering quality. We highlight two points: 
1) Better clustering indicates improved feature quality by eliminating the confounding demographic effects, such that the demographic features are independent of the rest causal identity features. Hence, images from the same attribute are distributed closely. 
2) Bias mitigation is not about removing the demographic attribute features, as they are indeed useful in identity differentiation. The bias lies in the over-dependence of the model on demographic-specific context. We formalize this intuition in Appendix.


\begin{figure}[t]
    \centering 
    \setlength{\abovecaptionskip}{8pt}
    \includegraphics[width=0.85\linewidth]{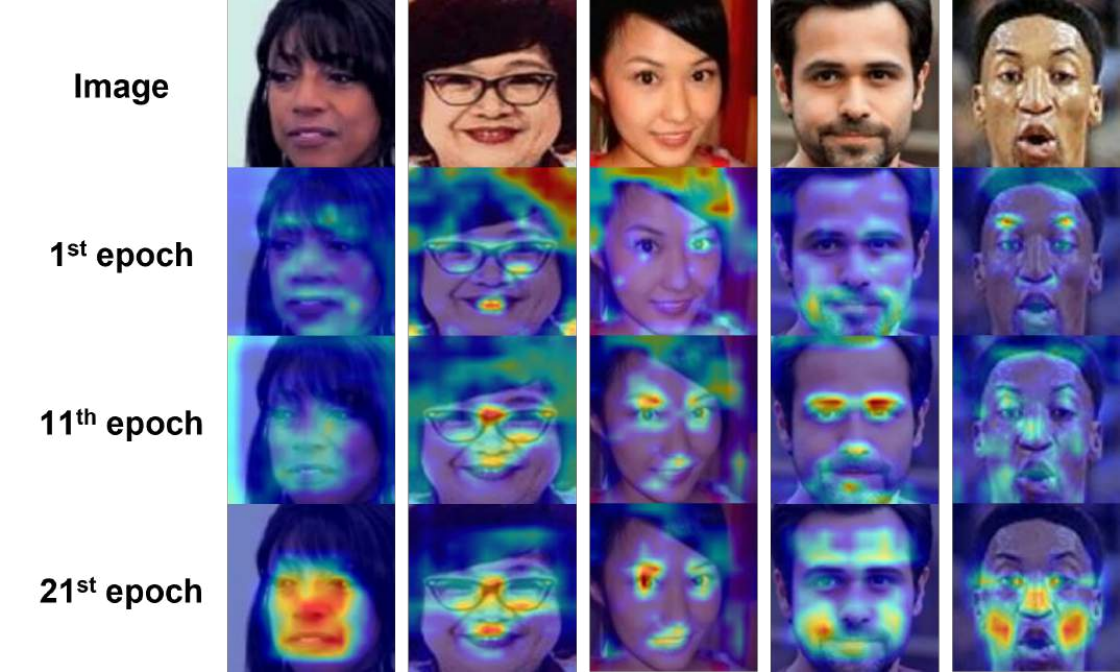} 
    \vspace{-0mm}
    \caption{Evolving attention maps during \textsc{Inv-Reg} learning (ResNet-50 backbone).}
    \label{fig:attention}
    \vspace{-4mm}
\end{figure}

\noindent\textbf{How does the attention map evolve in training?}
Figure~\ref{fig:attention} visualizes the Grad-CAM~\cite{selvaraju2017grad} in different training epochs.
We observe that, as the training progresses, the model learns to capture more causal features (\eg, facial characteristics) while eliminating spurious biases (\eg, hairstyle and facial hair). This further validates our invariant learning.

\section{Conclusion}

We presented a novel Invariant Regularization (\textsc{Inv-Reg}) for fair face recognition, which performs causal intervention to remove the confounding effect from the demographic attributes.
\textsc{Inv-Reg} iterates between learning a data partition as a self-annotated confounder, and pursuing invariant feature across the partition subsets to deconfound.
Through extensive evaluations on standard benchmarks, we show that \textsc{Inv-Reg} promotes fairness by improving the accuracy on various minority demographic groups without sacrificing that on the majority ones.
In particular, \textsc{Inv-Reg} is orthogonal to existing methods and can be freely combined with them to achieve improved performance.
In future work, we will seek other observational intervention algorithms for improved performance, and pursue an explainable model, \eg, by representation disentanglement~\cite{higgins2018towards}.

\section{Acknowledgements}

The authors would like to thank all reviewers and ACs for their constructive suggestions. 
Part of this research conducted at Nanyang Technological University is supported by the National Research Foundations, Singapore under its AI Singapore Programme (AISG Award No.: AISG2-RP-2021-022).
{\small
\bibliographystyle{ieee_fullname}
\bibliography{egbib}
}

\end{document}